# Futurity as Infrastructure: A Techno-Philosophical Interpretation of the AI Lifecycle[*]

Mark Coté [1, 2, †] and Susana Aires[1, †]

[1] *King's College London, London, United Kingdom*

**Abstract**
This paper argues that a techno-philosophical reading of the EU AI Act can offer fresh insight into the long-term dynamics of data within AI systems—specifically, how the lifecycle from data ingestion to model deployment generates recursive value chains that challenge existing regulatory frameworks for Responsible AI. We introduce a new conceptual tool to critically frame the AI pipeline, from data, training regimes, deep learning architectures, feature stores, and transfer learning processes. Drawing on cross-disciplinary methods, we develop a technically grounded and philosophically coherent analysis of regulatory blind spots. Our central claim is that what remains absent from contemporary AI policymaking is an account of the *dynamic of becoming* that underpins both the technical operation and economic logic of AI systems. To address this, we advance a formal reading of AI inspired by Gilbert Simondon's philosophy of technology, reworking his concept of *individuation*—a processual, non-static ontology—to model AI's developmental lifecycle. We distinguish three phases: i. the *pre-individual milieu*, where data, architectures, and parameters exist as latent potentials; ii. the process of *individuation*, where model coherence emerges through training, tuning, and integration; and iii. the *individuated AI*, which retains residual pre-individuality—ongoing capacities for adaptation, retraining, and cross-domain transfer. To translate these ideas into more applied terms, we introduce the concept of *futurity:* the self-reinforcing lifecycle of AI, in which increased data availability enhances model performance, deepens personalisation, and enables new domains of application. Futurity highlights the recursively generative, non-rivalrous nature of data in deep learning systems, underpinned by infrastructures like feature stores that enable real-time feedback, adaptation, and temporal recursion. Our intervention foregrounds the escalating power asymmetries at this critical juncture in history, particularly the tech oligarchy whose infrastructures of data capture, model training, and deployment concentrate value and decision-making power. We argue that the challenge of AI misalignment must be understood in light of these recursive value chains, and that effective regulation must account for the infrastructural and temporal dynamics of AI becoming. Our paper makes a number of regulatory proposals, including Lifecycle-based audit regimes, Temporal traceability, Feedback accountability, Recursion transparency, and the Right to Contest Recursive Reuse, measures that seek to reassert agency over futurity.

**Keywords**
Responsible AI, EU AI Act, Futurity, Technicity, Individuation, Google AI stack, AI value chain

## 1. Introduction

This paper argues that a techno-philosophical reading of the EU AI Act can offer fresh insight into the long-term dynamics of data within AI systems—specifically, how the lifecycle from data ingestion to model deployment generates recursive value chains that challenge existing regulatory frameworks for Responsible AI. We introduce new conceptual tools to critically frame technical objects in the AI pipeline, from data, training regimes, deep learning architectures, and transfer learning processes. Drawing on cross-disciplinary methods, we develop a technically grounded and theoretically sophisticated analysis of regulatory blind spots.

Our central claim is that what remains absent from contemporary AI policymaking is an account of the *dynamic of becoming* that underpins both the technical operation and economic logic of AI

---

systems. To address this, we advance a formal reading of AI inspired by Gilbert Simondon's philosophy of technology, reworking his concept of *individuation*—a processual, non-static ontology—to model AI's developmental lifecycle. We distinguish three phases: i. the *pre-individual milieu*, where data, architectures, and parameters exist as latent potentials; ii. the process of *individuation*, where model coherence emerges through training, tuning, and integration; and iii. the *individuated AI*, which retains residual pre-individuality—ongoing capacities for adaptation, retraining, and cross-domain transfer. We theorise the dynamic becoming of AI by introducing the complimentary concept of *technicity* which designates an excess or potential for new functionality present in all technical objects.

We apply this techno-philosophical framework to a concrete system, i.e., the Google AI stack, to trace the becoming of AI across seven interconnected stages—from data generation and capture to personalised inference and recursive feedback. At each stage, we examine both technical operations and their philosophical implications. We argue that data in this context is not a passive input but part of a dynamic, self-reinforcing system—a recursive infrastructure made possible by the non-rivalrous nature of data and its excludability within Google's proprietary pipeline. It is through this recursive architecture that we introduce the concept of *futurity*. Futurity describes how past user interactions and present system behaviours are recursively leveraged to refine predictions, personalise outputs, and extend model capabilities into new domains. It foregrounds the techno-economic logic underpinning AI: the continuous generation of value through temporally structured feedback loops. By mapping how Google's AI stack enacts this logic, we show how *the temporal logic of futurity* becomes material—operationalised not as an abstract potential but as an infrastructural condition of AI development and deployment.

We conclude by demonstrating how a techno-philosophical perspective—grounded in the concepts of individuation, technicity, and futurity—can generate new insights for regulatory design, particularly in relation to the EU AI Act. Specifically, we identify three interrelated blind spots. First, the Act does not adequately address the temporal infrastructures that drive ongoing and long-term model transformation; it regulates static systems rather than recursive ones. Second, while lifecycle obligations are acknowledged for high-risk systems, the Act lacks a robust framework for evolutionary governance—mechanisms that can track and audit systems as they adapt, personalise, and reconfigure post-deployment. Third, and most critically, the Act is silent on the political economy of value extraction: it offers no tools to address the structural asymmetries through which large platforms accumulate predictive capital by enclosing user interaction within closed-loop infrastructures. Our intervention foregrounds these escalating asymmetries—not as a failure of technical alignment, but as a consequence of the recursive, infrastructural dynamics of AI becoming. We argue that effective AI governance must move beyond static compliance and risk categorisation toward a temporal, infrastructural mode of regulation—one that can track how systems evolve, who benefits from their transformations, and how value is redistributed across time.

## 2. A Techno-Philosophical Reading of AI

Our conceptual reframing of current regulatory initiatives comes from a techno-philosophical reading of AI inspired by the work of Gilbert Simondon (1924-1989). Simondon was a French philosopher of technology concerned with understanding the nature of technical objects and their role in human life. Instead of succumbing to technophobic or technophilic views of technology, Simondon grounded his theorisation in a rigorous study of the technical functioning and evolution of specific objects, from tools to industrial and information machines. Albeit marginal beyond the French intellectual scene of his time, Simondon's work has been the object of growing attention in recent years among scholars aiming to theorise data and AI (Aires, 2024; Christen and Fabro, 2019; Coté and Pybus, 2016). He authored two major interrelated concepts: the theory of individuation put forward in *Individuation in light of notions of form and information* (2020) and, the theory of technology advanced in *On the mode of existence of technical objects* (2017). From these core tenets, we draw 1) the analysis of the *process of becoming* advanced in individuation theory and 2) the privileging of *technicity* underpinning Simondon's theory of technology. We posit that this techno-philosophical reading of AI can offer a novel understanding of Responsible AI and prompt a reappraisal of existing regulatory approaches.

Individuation is the cornerstone of Simondon's philosophy. It designates the *process of becoming*, a processual and relational ontology applicable to humans and technical objects alike (Simondon, 2017, 2020). Why we find Simondon so apposite for rethinking AI is that he does not limit 'individuation' or the 'individual' to humans, but rather posits becoming across myriad domains, from the physical to the biological to the social, and crucially, to the technical. Here we put forward individuation to explain the *becoming of AI* as it offers a processual understanding of AI as something that undergoes continual change after model deployment, not as fixed and stable entities. Individuation offers an innovative conceptual map of the AI lifecycle across three phases (See Figure 1). First is the *pre-individual state*, in which components like data, model architecture, training objectives and parameters exist as unstructured potential. Second is the *phase of individuation* proper which operationally coheres these components through training, tuning and system integration. Third we arrive at the *individuated AI system* - a functioning model embedded in application context. Crucially, this is not the terminal stage but rather a temporary resolution. The individuated AI system retains residual potential for further adaptation, reconfiguration, and transfer. This ongoing capacity for transformation marks the temporal openness we refer to as futurity, developed further in Section 3. For now, we put forward individuation as a conceptual language for understanding AI systems not as static entities, but as evolving infrastructures shaped by recursive feedback, temporal depth, and infrastructural becoming.

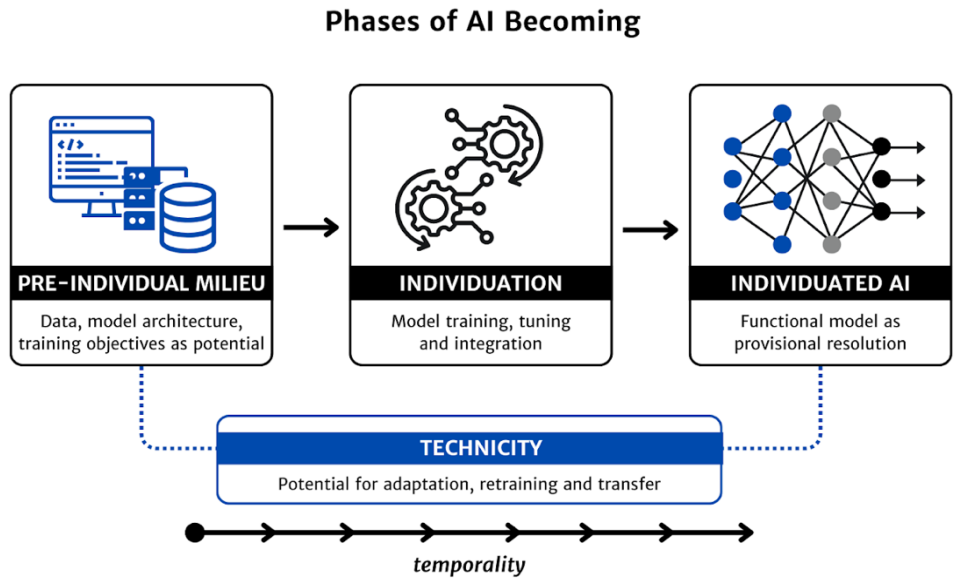

**Figure 1:** The three phases of AI becoming: 1) pre-individual milieu, where technical elements exist as potential; 2) the phase of individuation, where elements are put into relation to foster the becoming of the AI model, i.e., the emergence of a coherent schema of functioning, individuating model parameters through training, tuning and integration; 3) the individuated AI model emerges as a functional unit and a provisional resolution. Technicity - the engine of AI becoming - is present throughout as residue of potential for the temporal adaptation, retraining, transfer and transformation of data and model functionality.

Individuation does not ask 'what is', rather, 'how it came about', and does so in processual and relational terms, with a focus on transformative potential. The moment there is relation there is already a "system of individuation", wherein the elements-in-relation resonate with each other and trigger the individuation of the system, i.e., its differentiation or phase-shift (Simondon, 2020: 8). For example, we can consider model training as an individuating moment wherein relationality between a training dataset and a model architecture trigger the becoming of a functional AI model (Aires, 2024). This transformative potential of relation that underpins individuation ensues from the residue of potential present in all individuals and capable of triggering further individuations. Insofar as "individuation is an event and an operation within a reality that is richer than the individual that results from it" (Simondon, 2020: 53), it presupposes a domain rich in latent potentials which triggers individuation. Simondon calls this the *pre-individual reality*. The pre-individual does not correspond to a transcendental essence of all things but rather recognises the metastable condition of individuals, a contextual reality shared by the technical and the human. By positing *metastability* as the primary condition of individuals, Simondon criticises the view that beings move towards stability, towards a complete and final individual, which the philosopher regards as deprived of fecundity (Simondon, 2020). In contrast, the metastable denotes the provisional nature of individuals, acknowledging the potential and capacity for differentiation that they carry, i.e., *the pre-individual charge*, and that can, under specific conditions, be actualised and result in the emergence of a new structuration (e.g., the technical object).

Technicity is our other concept drawn from Simondon, which he puts forward in the theory of technology by prioritising the *technicity* of technical objects – their "mode of being" (Simondon, 2014: 82). The mode of being of technical objects is in their processual functioning, in Simondonian terms, the evolution of technicity. This requires a close study of the schemas of technical objects and how their functionality evolves. This examination is twofold: first zooming into the elemental components comprising a given technical operation, and second, in considering how these elements can be differently articulated to concretise novel schemas of functioning (Simondon, 2017). Technicity is the motor of Simondon's theory of technology, a quasi-genetic account of the incorporation and transformation of functionality at work in the invention of novel technical objects: we invent by drawing on existing technical elements and combining their functionality. For example, the ImageNet dataset has long been used to train and benchmark several state-of-the-art models. The *technicity* of the ImageNet dataset acts as part of the pre-individual reality of a plurality of AI models. As we shall demonstrate through the concrete example of Google's AI stack, we can think of these existing technical elements, e.g., data, model architecture, compute, among others, as the pre-individual milieu that will enable the concretisation of new AIs.

What is crucial about technicity is that it redirects our attention to AI by supplanting an understanding of technical objects grounded on *uses* to focus on their *functionality* that can positively contribute to ongoing and future regulatory initiatives. Technicity allows us to move beyond the limited uses that we give to technologies at a given point in time, towards a more perennial understanding of their schemas of functioning. This perennial understanding of functionality is in fact critical to address the highly recursive reality of AI, wherein technical elements have the potential for constantly being actualised and recombined to form novel, unanticipated sociotechnical realities and uses. We favour this theoretical frame as it adds a temporal dimension to the technical which is of utmost importance to address the long-term value chain of AI: technicity is mutable functionality that can be transformed through technical action, spanning the past, present and, crucially, the future.

This takes us back to individuation which can help us single out the unique challenges posed by AI technologies. As we will see, individuation emphasises how the dynamism of AI becoming energises the long-term value chain of AI, and how the technicity *qua* potentiality of data and system is constantly being reused and repurposed to form new AIs without ever being exhausted.

It is worth restating the paramount technical role of data for AI, underpinning the pre-individual reality of contemporary AI models, which ingest large amounts of data to extract patterns of functionality. Data harbours a potentiality that stems directly from our everyday lives, capturing micro and latent sensibilities on a vast scale, with the aim of representing, quantifying and rendering actionable multiple domains of our lifeworld. As processes of datafication occur at an unprecedented scale, data are not only an ever-growing asset – as primary and derived data – but a domain of potentiality that is never exhausted in the individuation of AI, for data can be infinitely mined (see

Non-rivalry below). What follows from data, as a domain of pre-individual potentials, is the capacity to foster the individuation AI models through training, wherein the recursive processing of data edifies model functioning by individuating its parameters, with the technicity of data being 'transferred', recombined and transformed into model functionality. The data and model architecture that previously existed as potential, give rise to a new realm of functionality that is highly generative and can itself ground a new lifecycle of repurposing, as is the case of foundation models.

Analogously, the rollout of several foundation models in the past few years, including generative and multimodal models, attests to the fact that technicity is not simply given functionality but the potential of functionality to form the basis of new AIs. The fact that the EU AI Act risked being outdated before coming into effect, due to foundation models challenging the limited intended uses listed in the Act, attests to this fact. Technicity is a residue of potential ever-present in individuated technical objects–including the data and models comprising AI–which can always foster new functionality. Moreover, this becoming is always contingent on the particularities of that technical object, for example, with foundation models which can only be known through *ex-post* close study. We will present this as a pivot point for the EU AI Act to shift from a solely risk-based approach – which reduces the regulatory framework to specific use-cases – to a processual orientation capable of tackling the technicity of AI.

From data to model functionality, the becoming of AI ultimately entails the following: 1) we generate data, 2) data enter model architecture fostering the becoming of new AIs, 3) these models are deployed in AI systems and fine-tuned through user interactions, 4) AI systems generate new data that is fed back into life, shaping experiential reality and action, and 5) models can be repurposed into new domains repeating the cycle. This data-system individuation—which is inherently a techno-human co-individuation—is what is at stake in the becoming of AI, not just static technical objects. As we shall demonstrate through the case-study of Google's AI stack, this human-technical co-individuation entails not simply the recursive becoming of the model 'in itself' through user-model interaction, but underscores how technicity cuts across multiple dimensions of life which are also economic and political in nature.

Summarising, we offer this philosophical frame to make visible the dynamic of AI becoming underpinned by technicity in contrast to static technical objects with fixed functionality and risks. Our techno-philosophical model gives an innovative perspective on the broader lifecycle of AI, framing a technically-grounded appraisal and regulatory rethink that brings into critical focus not just the lack of accountability but the alarming escalation of political and economic power disparities. Individuation and technicity let us see the temporal dimension in AI and the accelerating power these multi-functional models afford to the tech oligarchs which possess them.

## 3. Case Study: the Google AI Stack as Futurity Infrastructure

Now we map the recursive lifecycle of data within the Google AI stack, building on the socio-techno method of the SDK Data Audit (Pybus and Coté 2024, Pybus and Mir 2025, Coté and Pybus, 2021). We illustrate how the ostensibly linear AI pipeline constitutes an expansive closed-loop system of predictive generation, materialising what we term *futurity as infrastructure*. Below we trace seven interconnected stages, foregrounding moments of becoming (individuation) and implications for temporality, agency and value. Before turning to the case study, we introduce three foundational concepts that frame our analysis: *non-rivalry*, *excludability*, and *futurity*.

*Non-rivalry*, in economic terms, denotes data's capacity for infinite reuse without depletion. Unlike physical goods, data can be replicated and deployed simultaneously across multiple applications, domains, and systems while maintaining its full utility. For example, the same user interaction data that trains a recommendation model can simultaneously refine a language model or drive real-time moderation–all at once.

*Excludability* refers to the technical and legal infrastructures that restrict access and generate enclosure. While data may be non-rivalrous, it becomes excludable through platform architectures–such as closed APIs, proprietary SDKs, and paywalled services. This combination of infinite reuse

(non-rivalry) and controlled access (excludability) underpins the political economy of data capitalism: data can generate compound value over time, but only for those who control and govern the infrastructure through which it flows.

*Futurity*, as developed by Coté (Forthcoming), refers to *the monetisable orchestration of time in data-driven AI systems.* It captures how these systems transform past interactions and present behaviours into predictive outputs that preconfigure future actions. *Data*, in this view, *is not merely a byproduct of use, but the substrate of a generative feedback loop.* Once captured and standardised–through infrastructures like feature stores–data is continuously transformed into model refinements, actionable insights, and anticipatory interventions. In this loop, AI systems do not merely respond to the world—they act on it pre-emptively, producing value by folding historical traces into the conditions of future behaviour.

The conceptual dimensions of *futurity* can be summarised as follows. First, *data as temporal experience*: data encodes lived user interactions across multiple temporal scales; models are trained on the past to act on the present and modulate the future. Second, *recursive feedback*: each user interaction generates new data that refines future predictions, eliciting further interaction. Third, *model development*: the infrastructure of futurity materialises in model evolution–through fine-tuning, domain adaptation, and transfer learning. Fourth, *actionable prediction*: beyond forecasting, AI systems generate personalised recommendations, nudges, flags, and decisions that actively modulate user behaviour. Fifth, *monetisation*: these outputs are economically significant, underpinning the platform's business model and reinforcing data value asymmetries. Monetisation serves as a dynamic and dominant logic articulating all four dimensions—integrating recursive feedback, inference generation, actionable prediction, and continuous model training. Thus, monetisation drives recursivity within the dynamic AI lifecycle (See Figure 2).

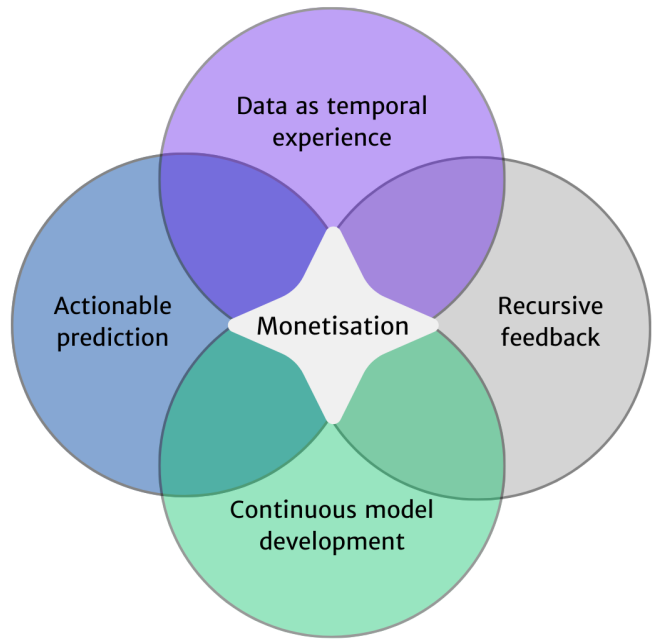

**Figure 2:** The conceptual dimensions of Futurity.

To make this concept of futurity operational, we present a seven-stage case study of the Google AI stack (See Figure 3). This lifecycle illustrates how user interaction becomes predictive capital through recursive feedback loops, infrastructural enclosure, and temporally extended model refinement. Each stage is presented in two layers: first, a technical account of the data pipeline; and second, a techno-philosophical reading of its systemic implications.

## 3.1. App to Firebase — Data Generation and Capture

*Technical Summary*: The lifecycle begins with the user interacting with a mobile application. Actions such as logging a workout, dismissing a notification, or entering a free-text note are captured by the Firebase SDK embedded in the app. These interactions are recorded along with device metadata and contextual information such as time, location, or connectivity state. This event-level data is stored in Firebase and becomes the foundational substrate for downstream machine learning processes.

*Techno-Philosophical Reflection*: At this stage, user interaction becomes a site of data extraction. The user's lived experience is rendered into machine-readable form—what Simondon might call the entry of pre-individual potential into a system of individuation. Data is not simply generated; it is captured and formalised, signalling the initial move from embodied behaviour to infrastructural trace. This is the moment when human action is made computationally actionable.

## 3.2. Firebase to BigQuery — Data Structuring and Preprocessing

*Technical Summary:* Raw event data from Firebase flows into two pipelines: i. a batch path via BigQuery for historical structuring and training, and ii. a real-time path via Pub/Sub and Dataflow for feature transformation and contextual inference. In BigQuery, data is cleaned, standardised, and shaped into structured tables to support model training and generalisation. In parallel, Dataflow engineers features from raw event data and writes them to the Feature Store for low-latency inference. In addition to supporting scalable storage, access control, and compliance, these dual paths enable the system to meet the demands of both retrospective learning and real-time prediction—two temporalities fundamental to AI's recursive becoming.

*Techno-Philosophical Reflection:* This phase marks the transition from raw behavioural traces to structured, repeatable, and monetisable information. The data is both non-rivalrous and excludable: it can be reused indefinitely but remains under the platform's control. It is here that data acquires its infrastructural character—no longer a single-use artifact, but a reusable input into recursive model improvement. From a Simondonian perspective, this stage enhances the pre-individual potential of data, refining its technicity to support downstream processing. We can also discern a temporal orchestration of individuation: the training pipeline (BigQuery) forms generalised structures—the 'long memory' of the model—while the inference pipeline (Feature Store) modulates situated action, serving as the system's immediate 'intuition'.

## 3.3. BigQuery to TFX — Model Training and Orchestration

*Technical Summary:* Structured training data from BigQuery flows into TensorFlow Extended (TFX), Google's end-to-end machine learning platform. TFX orchestrates the batch training pipeline, including schema validation, feature transformation, and model development. A general-purpose model is trained on population-level data to identify broad behavioural patterns, preferences, and user clusters. This training process encodes the system's long memory, producing functional predictive structures that can later be adapted and personalised during inference. TFX ensures standardisation, reproducibility, and consistency across training environments.

*Techno-Philosophical Reflection:* Here individuation occurs through transduction: the system processes and integrates the pre-individual elements—data, model architecture, training objectives—into a coherent and operational model. The system moves from potential to functioning structure, forming a predictive entity capable of generalising across a population. It is no longer data alone, but data shaped by learning objectives and model logic into a structure of becoming.

## 3.4. TFX to Vertex AI — Deployment and Personalisation Infrastructure

*Technical Summary:* Following training, a general-purpose model is exported, packaged, and deployed via Vertex AI, Google's managed platform for serving and scaling AI applications. This deployment enables real-time inferencing and supports segmentation into behavioural personas derived from training on population-level patterns. The *feature store*—central to this deployment—is initialised with precomputed contextual and population-level features derived during training, setting the conditions for recursive adaptation. The feature store supports a mutable prediction layer which dynamically combines initial features with live user inputs such as location or recent activity—to generate tailored responses in real time.

*Techno-Philosophical Reflection:* At this stage, the model ceases to be merely latent potential and becomes individuated in the Simondonian sense: it functions as a coherent, embedded technical object within a live system. The model not only processes inputs and produces outputs, but also sustains internal consistency and responsiveness over time. Personalisation becomes the visible manifestation of this individuation: the system acts on users in ways shaped by both historical population training and emergent segmentation logics. It is not a static artefact, but a modulating situated system—operational, adaptive, and relational.

### 3.5. Personalised Inference — Real-Time Prediction and Granular Futurity

*Technical Summary:* At the moment of user interaction, the deployed model queries the feature store to retrieve the user's most recent contextual signals—such as history, recent actions, location, time-of-day, or app-specific activity. These features are combined with the model's pre-trained weights and behavioural segments to generate a low-latency, best-guess prediction tailored to the user's current situation. These real-time features are served through a dedicated inference pipeline—typically using Pub/Sub and Dataflow—optimised for speed and responsiveness. The feature store acts as the hinge between the model's static knowledge and the user's dynamic context, enabling inference to be situationally aware. Crucially, this occurs *before the user acts*, making inference a predictive intervention rather than a reactive computation.

*Techno-Philosophical Reflection:* This is the most immediate instantiation of *futurity*. Inference becomes a site of temporal orchestration: the model mobilises the past to shape the present and condition the future. This is not merely prediction—it is modulation. The technicity of the model extends into the user's world, narrowing their field of possible actions through nudges, notifications, or recommendations. In this way, the model co-produces its own training substrate, as the inference directly influences the user's behaviour, which is then recaptured as feedback. Personalised inference is not an endpoint; it is a hinge. It connects historicised learning with anticipatory action, and in doing so, initiates the recursive loop that characterises AI becoming.

### 3.6. Adaptive Feedback Loops — Recursive Learning from User Response

*Technical Summary:* This occurs after the user has acted, either engaging, ignoring, or resisting the inference. These responses are captured via Firebase and structured into new feature values—such as recent engagement frequency, time since last interaction, or inferred user intent. These features are then written back into the feature store, where they become immediately available for future inferences. In this way, the system adapts to user behaviour in near real-time, refining its outputs without the need for full model retraining. The feature store thus serves as the temporal hinge of adaptation: personalisation deepens as the system learns not from fixed ground truths, but from lived interaction–and renders those data points actionable for the next prediction cycle.

*Techno-Philosophical Reflection:* Here the recursive infrastructure of *futurity* fully takes shape. The system does not simply predict; it learns from the efficacy of its own predictions. The system adapts to itself over time, creating a loop of co-individuation: the model evolves as the user responds to its outputs, and the user's future experience is shaped by the model's adaptive recalibration. This is *becoming* in action—a temporally extended infrastructure that folds the user into its own developmental logic. The clear boundary between model and data dissolves; inference becomes an engine of ongoing transformation.

### 3.7. Reintegration — Recursive Infrastructure and Value Accumulation

*Technical Summary:* The feedback data captured in Stage 6 re-enters the pipeline in two ways: as structured features, immediately available for inference via the feature store; and as new training material for batch-based model retraining. This closes the loop between user behaviour and model development, enabling continuous refinement, deeper personalisation, and expansion of predictive capacity. Over time, the system compounds value—each cycle increases its granularity and scope of intelligibility. This recursive becoming is embedded within Google's vertically integrated infrastructure: Firebase captures interaction data; BigQuery processes and structures it; TFX orchestrates retraining; and Vertex AI serves updated models and supports real-time inferencing. The system's futurity is encoded in this stack—where past actions prefigure future outputs, and present behaviour shapes the conditions of the next prediction. Stage 7 thus completes the temporal arc of AI individuation: not a static loop, but a recursively generative infrastructure in continuous becoming.

*Techno-Philosophical Reflection:* This final stage exemplifies the political economy of *futurity*. The system functions as a closed loop, where value is extracted from time—past behaviours become future capital. User agency is enclosed within a recursive infrastructure in which each gesture fuels future prediction and monetisation. Data is non-rivalrous—it can be reused endlessly—and excludable—it remains locked inside proprietary systems. The outcome is a form of recursive enclosure: a predictive infrastructure that grows smarter and more asymmetrical with every cycle, where user behaviour is mined as predictive capital in perpetuity.

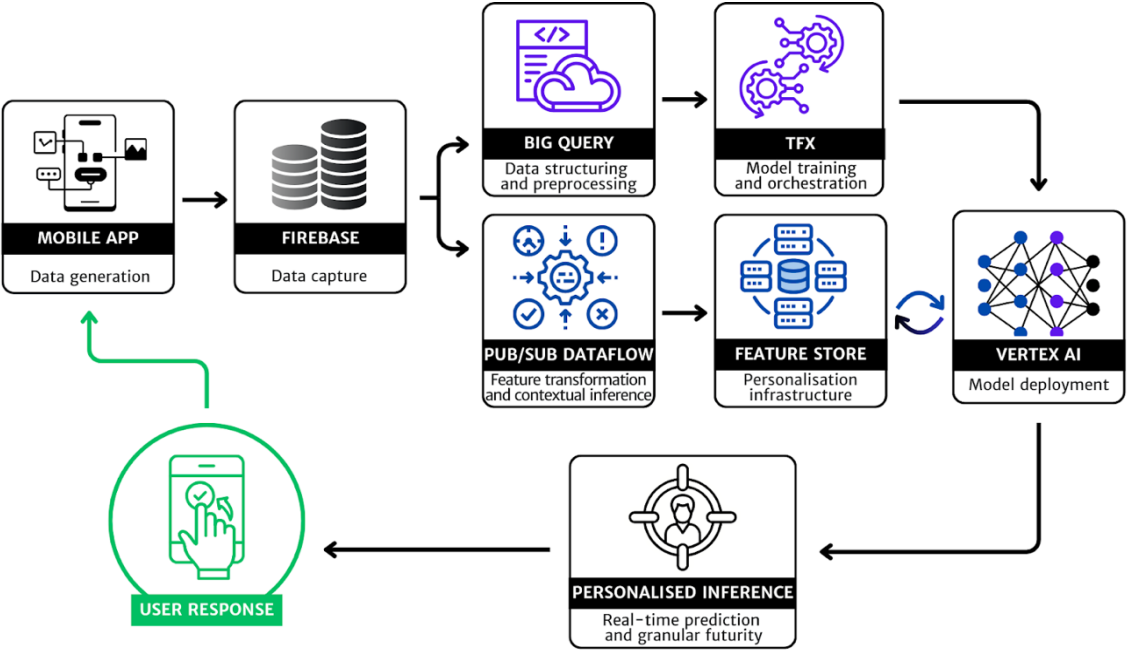

**Figure 3:** The Google AI Stack as a futurity infrastructure. The AI lifecycle begins with 1) User data being generated and captured by Firebase. 2) Data stored in Firebase flows into two pipelines: a batch-based training path via BigQuery and a real-time inference path via Pub/Sub and Dataflow. 3) Structured data from BigQuery is used for model training and orchestration in TensorFlow Extended (TFX) while features engineered by Dataflow are written into the Feature Store for low-latency inference. 4) The trained model is exported, packaged, and deployed via Vertex AI. 5) At the moment of user interaction, the deployed model queries the feature store to retrieve the user's most recent contextual signals, generating a personalised inference before the user acts. 6) Once the user acts, user responses are captured via Firebase and structured into new feature values which become immediately available for real-time prediction via the feature store. At the same time, user responses become new training material for model retrospective learning. 7) The closed loop between user behaviour and model deployment enables the reintegration of data into the Google AI Stack, which becomes an infrastructure in recursive becoming, with deepening predictive accuracy and growing economic capital.

Together, these seven stages form an infrastructure of futurity—a system capable of not only predicting behaviour, but of recursively shaping it, personalising outputs in the present, and learning from interactions to generate value in the future. Together, they map the individuation of AI not as a discrete act, but as a recursive process linking data capture, model training, system deployment, inference, and adaptive feedback into a continuous infrastructure of becoming. This orchestration relies on two interlocking temporal pipelines: a batch-based training path, which aggregates historical data for generalisation, and a real-time inference path, in which the feature store acts as a temporal hinge—connecting the model's general knowledge to the user's dynamic context. Inference and feedback do not represent the end of AI; they form a generative loop in which past behaviour becomes the substrate for future intervention. Through this lens, the Google AI stack exemplifies how futurity operates not as a future event, but as an infrastructural condition. Notably, this is as a closed, vertically integrated system in which user data is recursively transformed into capital–both predictive and financial. It is within this loop that personalisation deepens, value accumulates, and user agency is progressively directed by the logic of the model. This infrastructure operationalises and monetises futurity.

## 4. Futurity Governance Mechanisms: A Techno-Philosophical Reading of the EU AI Act

Our techno-philosophical reading of the EU AI Act is intended not merely as critique, but as a speculative contribution to a shared imaginary of what AI governance could look like in a post-AI Act landscape. Central to this reimagining is the insight that the terrain on which AI systems operate is dynamic, not static—that becoming and futurity are fundamental features of AI systems and should be integrated into their regulation. We identify three major blind spots in the Act from this perspective (See Figure 4).

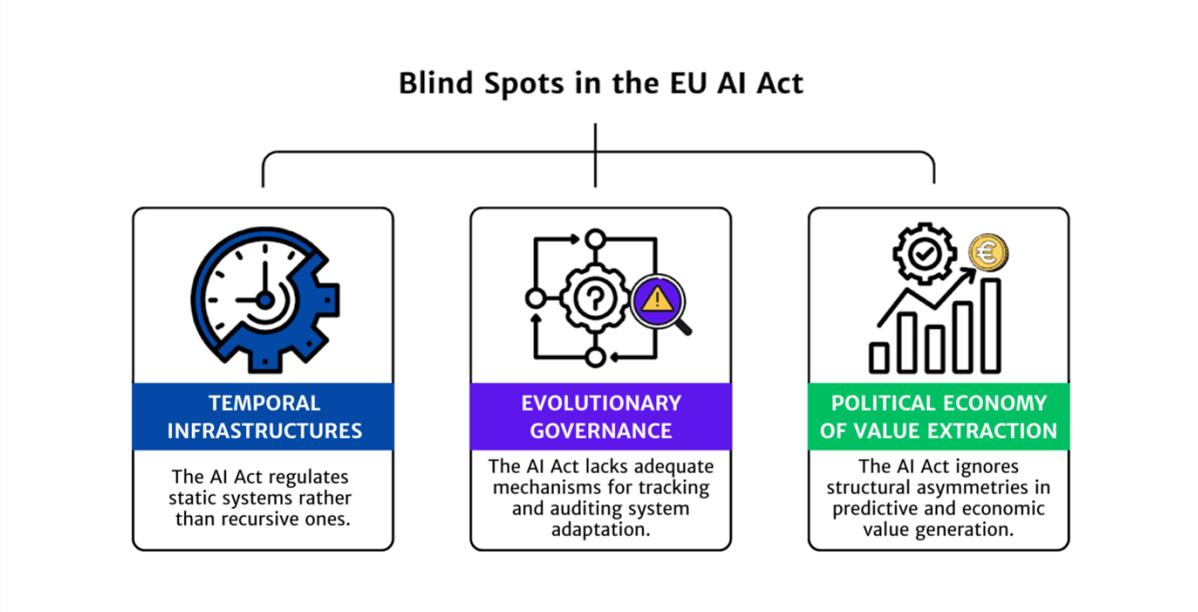

**Figure 4:** Blind spots in the EU AI Act.

First and fundamentally, the AI Act misses AI as being temporally recursive. The EU AI Act treats AI systems as bounded, classifiable objects, assessed for risk categories based on their immediate functionality. But AI systems today are temporally recursive: they evolve through feedback loops, learn from user interactions, and refine themselves post-deployment. Futurity foregrounds this reality, framing AI as an infrastructure of continuously becoming—shaped by historical data and generating new data to shape future behaviours. In other words, the Act regulates outputs (risk, performance), but not the processes of becoming that recursively generate those outputs over time.

This leaves futurity as infrastructure—the temporal infrastructures that drive long-term model transformation—in a blindspot.

Second, a techno-philosophical interpretation of the becoming of AI systems brings critical attention to the Act's *ex ante* regulatory logic: systems are assessed before being placed on the market. *Ex ante*—Latin for beforehand—is codified in Article 9 Risk management System and Article 10 Data and Data Governance, which require providers to ensure high-quality training data, design controls, and conformity assessments *prior* to deployment. *Ex ante* accountability frameworks assume risk can be assessed and mitigated in advance, a logic that breaks down in the face of recursive AI systems that can adapt and evolve with every user interaction. Futurity, by contrast, demands lifecycle-based regulation, and shifts the focus from initial model evaluation to ongoing, post-deployment governance.[1]

The precise nature of temporal governance mechanisms exceeds the scope of this paper. Yet if regulation is to remain relevant in the face of recursive, evolving AI systems, *futurity-proofing* must become a guiding principle of governance design. At minimum, this entails developing *lifecycle-based audit regimes* that move beyond static compliance checks and instead address the evolutionary trajectories of AI systems.

Here we briefly outline several conceptual pillars that could guide such an approach. *Temporal Traceability* would track how predictions and model outputs change over time, and what data transformations or interactions have shaped those changes. *Feedback Accountability* makes visible how user interactions—clicks, dismissals, completions—feed back into model updates or personalisation strategies, asking who is being predicted *from*, and who benefits from those predictions? *Recursion Transparency* identifies which inputs (users, datasets, sources) contribute to which model outputs, at what stages of training or inference. This would enable users, auditors, and regulators to trace how past data *re-enters* the system. Finally, there is the *Right to Contest Recursive Reuse* which would afford individuals the ability to challenge or opt out of **recursive data reuse**—that is, the use of their behavioural data to continuously refine, personalise, or monetise AI outputs, often without their knowledge or consent. This would shift regulation from simply protecting users from harm, to **reasserting agency over futurity**—to recognise data not only as personal, but as temporal, infrastructural, and political.

Finally, there is the giant blindspot to the asymmetries of value capture, that is, whereas the Act seeks to mitigate harm, it does not challenge how AI systems *extract value from time,* how predictive infrastructures enclose user agency and distribute benefits unevenly. As we illustrated in our case study, the Google AI Stack can also be considered a platform of extractive value generation, wherein every user interaction becomes a training substrate for increasingly personalised, commercialised prediction. Yet this infrastructural dynamic remains wholly outside the current scope of AI regulation.

## 5. Conclusion

We conclude not in despair at the hundreds of billions funnelled into the hands of tech oligarchs via these self-reinforcing systems. Instead, we offer a few speculative but actionable proposals to reorient the flow of AI futurity toward public value. First, we propose temporal value disclosure—a regulatory mechanism requiring providers to report how much of their model performance is attributable to post-deployment user data, and to quantify the extent to which real-time feedback loops contribute to ongoing system optimisation. This would foreground the temporality of value generation that currently escapes visibility in regulatory reporting. Second, we call for infrastructure transparency requirements that would mandate clear reporting on data flows, from user input to model output, and provide visibility into access and control over key pipeline components such as Firebase, BigQuery, TFX, and Vertex AI. While proprietary claims may remain, such reporting would at least enable auditing bodies and civil society actors to observe how these systems operate recursively and at scale. Third, we propose the introduction of an AI windfall tax, which could support a public Futurity Value Redistribution Fund. This fund would be dedicated to strengthening public-sector AI capacity or building a federated AI Data Commons, making the Common European Data Spaces a reality. Importantly, the legal infrastructure for such an initiative already exists in the form of the Data Altruism provision within the Data Governance Act (Article 20). This could be leveraged to require large-scale AI providers to contribute a portion of their derivative embeddings, anonymised model outputs, or synthetic datasets—particularly those derived from public data—back into commons-based infrastructures. Such a system would be especially impactful in critical domains like health, education, and labour, where private-sector dominance risks eroding public knowledge and capacity.

---

[1] We are aware that the EU AI Act foresees the existence of a "post-market monitoring system" to be implemented by providers of high-risk AI systems, as per Article 72 (Regulation 2024/1689). Although the European Commission is due to provide a template for this purpose in 2026, we are worried that the emphasis on *ex ante* conformity assessment may result in monitoring systems becoming a box-ticking exercise with limited regulatory oversight. Moreover, the dismissal of recursive learning as a requirement for a new conformity assessment (Article 43), except in the event of "substantial modification", may hinder the fundamental aspect of lifecycle-based evaluation and governance (ibid).

We recognise that these proposals may sound ambitious, even utopian. But they are grounded in existing systems, conceptually coherent with the demands of temporally recursive AI governance and ultimately aligned with the public interest. When contrasted with the de-regulationist ideologies advanced by figures like Musk or Trump, which actively undermine the possibility of meaningful oversight, we argue that such proposals are not only timely—they are necessary.

## 6. Declaration on Generative AI

During the preparation of this work, the authors used Canva for all figures in order to: Generate images. After using this tool, the authors reviewed and edited the content as needed and take full responsibility for the publication's content.